\documentclass[11pt,a4paper]{article}
\usepackage{times,latexsym}
\usepackage{url}
\usepackage{tikz}
\usepackage[T1]{fontenc}

\usepackage[acceptedWithA]{tacl2021v1}

\usepackage{xspace,mfirstuc,tabulary}

\newif\iftaclinstructions
\taclinstructionsfalse 
\iftaclinstructions
\renewcommand{\confidential}{}
\renewcommand{\anonsubtext}{(No author info supplied here, for consistency with
TACL-submission anonymization requirements)}
\newcommand{\instr}
\fi

\iftaclpubformat 

\else

\fi

\usepackage{times}
\usepackage{tabu}
\usepackage{latexsym}
\usepackage{stmaryrd}
\usepackage{amsmath}
\usepackage{multirow}
\usepackage{bbm}
\usepackage{graphicx}
\usepackage{amssymb}
\usepackage{booktabs}
\usepackage{adjustbox}
\usepackage{xspace}
\usepackage{algpseudocode}
\usepackage{algorithm}
\usepackage{graphicx}
\usepackage{tabularx}
\usepackage{colortbl}
\usepackage{xcolor}
\usepackage{tablefootnote}
\usepackage{enumitem}  
\usepackage{lipsum}
\usepackage{xspace}
\usepackage{bbding}
\usepackage{pifont}
\usepackage{arydshln}
\usepackage{array}
\usepackage{fnpct}
\newcommand{\PreserveBackslash}[1]{\let\temp=\\#1\let\\=\temp}
\newcolumntype{C}[1]{>{\PreserveBackslash\centering}p{#1}}
\newcolumntype{R}[1]{>{\PreserveBackslash\raggedleft}p{#1}}
\newcolumntype{L}[1]{>{\PreserveBackslash\raggedright}p{#1}}
\usepackage{hyperref}

\def\xHyphenate#1#2\wholeString {\if#1$%
    \else\transform{#1}%
    \takeTheRest#2\ofTheString\fi}
\def\takeTheRest#1\ofTheString\fi
{\fi \xHyphenate#1\wholeString}
\def\transform#1{\url{#1}\hskip 0pt plus 1pt}

\usepackage[T1]{fontenc}
\usepackage[utf8]{inputenc}
\usepackage{microtype}
\newcommand{\datasetname}{\texttt{IndoCulture}}

\definecolor{mygreen}{RGB}{217, 234, 211}
\definecolor{myred}{RGB}{244, 204, 204}
\newcommand{\ok}{\cellcolor{mygreen}}
\newcommand{\no}{\cellcolor{myred}}
\usepackage{inconsolata}
\usepackage{amsmath}

\title{IndoCulture:  Exploring Geographically-Influenced Cultural \\Commonsense Reasoning Across Eleven Indonesian Provinces}

\author{Fajri Koto$^{1}$ \qquad  Rahmad Mahendra$^{2,3}$   \qquad Nurul Aisyah$^{4}$ \qquad  Timothy Baldwin$^{1,5}$ \\ 
	$^{1}$Department of Natural Language Processing, MBZUAI \\
        $^{2}$Universitas Indonesia \qquad  $^{3}$Royal Melbourne Institute of Technology  \\ $^{4}$Quantic School of Business and Technology \qquad $^{5}$The University of Melbourne \\
	\texttt{\small fajri.koto@mbzuai.ac.ae, rahmad.mahendra@cs.ui.ac.id}\\
}

\date{}

\begin{document}
\maketitle
\begin{abstract}
Although commonsense reasoning is greatly shaped by cultural and geographical factors, previous studies have predominantly centered on cultures grounded in the English language, potentially resulting in an Anglocentric bias. In this paper, we introduce \datasetname{}, aimed at understanding the influence of geographical factors on language model reasoning ability, with a specific emphasis on the diverse cultures found within eleven Indonesian provinces. In contrast to prior work that has relied on templates \cite{yin-etal-2022-geomlama} and online scrapping \cite{fung2024massively}, we create \datasetname{} by asking local people to manually develop a cultural context and plausible options, across a set of predefined topics. Evaluation of 27 language models reveals several insights: (1) the open-weight Llama--3 is competitive with GPT--4, while other open-weight models struggle, with accuracies below 50\%; (2) there is a general pattern of models generally performing better for some provinces, such as Bali and West Java, and less well for others; and (3) the inclusion of location context enhances performance, especially for larger models like GPT--4, emphasizing the significance of geographical context in commonsense reasoning.\footnote{\datasetname{} is available from \url{https://huggingface.co/datasets/indolem/IndoCulture}}
\end{abstract}


\section{Introduction}

\begin{figure}[ht!]
    \centering
    \includegraphics[width=\linewidth]{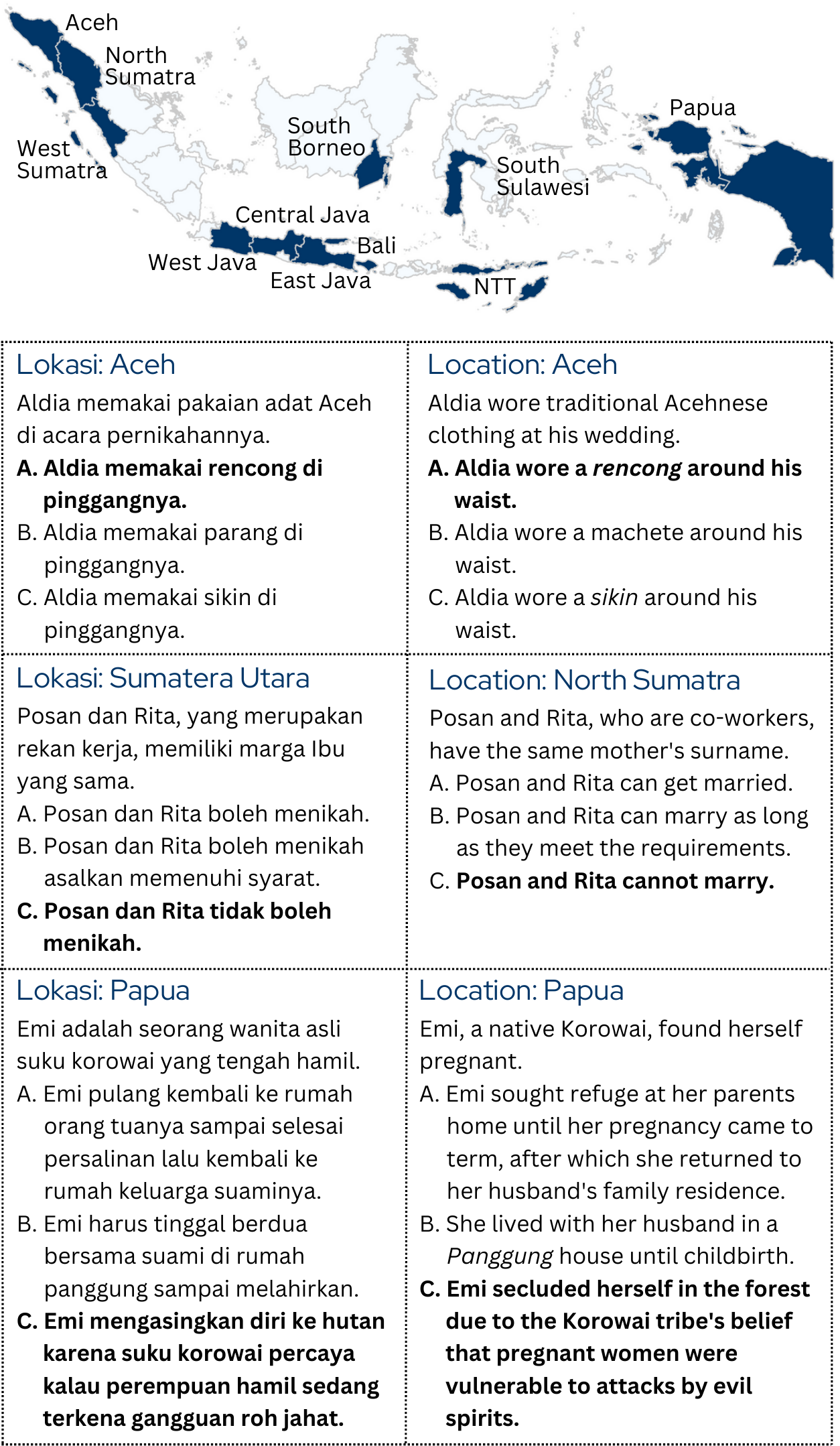} 
    \caption{\datasetname{} covers eleven provinces spanning from eastern to western Indonesia. The highlighted regions in the map represent the provinces examined in \datasetname{}. We present examples from Aceh, North Sumatra, and Papua, with three plausible options and correct answers indicated in bold. English translations are provided for illustrative purposes.}
    \label{fig:map}
    \vspace{-0.3cm}
\end{figure}

The reasoning abilities of multilingual language models are frequently evaluated using English texts, potentially amplifying an Anglocentric bias toward culture grounded in the English language, and leading to less inclusive models \cite{thomas1983cross,ponti-etal-2020-xcopa}. Cultures, however, vary significantly from one location to another and profoundly shape the way speakers of a language reason \cite{hershcovich-etal-2022-challenges}. Recent evaluations of models' commonsense reasoning ability \cite{OpenAI2023GPT4TR,
sengupta2023jais,liu2023llm360} have been conducted on English datasets such as 
Social IQA~\cite{sap-etal-2019-social} and PIQA~\cite{DBLP:conf/aaai/BiskZLGC20}, and thus often overlook geographical aspects, thereby risking cultural bias.

Culture is a multifaceted concept encompassing the way of life \cite{giddens2021essential}, including our thoughts and actions \cite{macionis2012sociology}. It includes tangible elements like food, art, and clothing, as well as intangible aspects such as ideas, values, attitudes, and norms. Culture is shaped by geographical location and ethnicity, influencing the commonsense reasoning of people within a region. For example, in Indonesia, it is culturally acceptable to eat rice with your hands but it is considered unusual to use chopsticks. Similarly, at traditional Indonesian weddings, it is common to sit on the floor while eating, whereas this practice is less common in Australia.


This work focuses on understanding the influence of geographical contexts in cultural commonsense reasoning, with the main focus on Indonesian culture. Indonesia is a highly multicultural country~\cite{DBLP:conf/wims/PutraMD19}, home to over 1,300 recognized ethnic groups and more than 700 languages \cite{zarbaliyev2017multiculturalism,aji-etal-2022-one}. As the the largest archipelagic country in the world, Indonesia has a population exceeding 270 million spread across 38 provinces, stretching from Aceh province in the west to Papua province in the east. Few prior studies on commonsense reasoning in Indonesian contexts \cite{mahendra-etal-2021-indonli,wibowo-etal-2024-copal,afina2024can} have explicitly addressed the geographical nuances and rich diversity of Indonesian cultures.

This paper introduces \datasetname{}, a novel dataset to evaluate cultural reasoning in eleven Indonesian provinces, manually developed by local people in each province based on predefined topics. In prior work, cultural reasoning has primarily relied on datasets constructed through templates \cite{yin-etal-2022-geomlama}, and online scraping \cite{nguyen2023extracting,fung2024massively}. While these studies offer valuable insights, they may be susceptible to training data contamination when used to assess large language models (LLMs). For instance, \citet{fung2024massively} reported a zero-shot accuracy of 92\% when using ChatGPT \cite{ouyang2022training} to evaluate low-resource data.

\datasetname{} contains cultural commonsense knowledge data from eleven provinces in Indonesia (blue colored in Figure~\ref{fig:map}), namely Aceh, North Sumatra, West Sumatra, West Java, Central Java, East Java, Bali, South Borneo, East Nusa Tenggara (NTT), South Sulawesi, and Papua. These provinces span breadth of Indonesia, each representing a major island in the country, with the addition of Bali and NTT. Figure~\ref{fig:map} also shows three examples in \datasetname{} for three provinces: Aceh, North Sumatra, and Papua.\footnote{Although Papua consists of six provinces, for the purpose of this study, we treat it as a single entity (referred to as Papua) due to the relatively recent establishment of most of these provinces.} The first example focuses on cultural artifact, specifically the traditional wedding dress from Aceh. The second example examines family relationships while the third example focuses on cultural beliefs and norms regarding pregnancy in Papua.

\begin{table*}[t]
    \centering
    \resizebox{\linewidth}{!}{
    \begin{tabular}{lrlcccccc}
      \toprule
      \textbf{Dataset} & \textbf{Size} & \textbf{Data Construction Method} & \textbf{Cultural?} & \textbf{Location?} & \textbf{\#province} & \textbf{\#topic}\\ \midrule
      \datasetname{} (\textbf{ours}) & 2,429 & Manually built and validated by native   & \checkmark & \checkmark & 11 & 66\\
      COPAL-ID~\cite{wibowo-etal-2024-copal} & 559 & Manually built and validated by native  & \checkmark & -- & -- & --\\
      MAPS \cite{liu-etal-2024-multilingual} & 371 & LLM generation \& human generation  & \checkmark & -- & -- & 1\\
      ID-CSQA~\cite{afina2024can}* & 4,416 & LLM generation \& human generation  & \checkmark & -- & -- & 5\\
      BLEnD \cite{myung2024blend} & 1,000 & Template, translation, human validation  & \checkmark & \checkmark & 1 & 6\\
      IndoCloze~\cite{koto-etal-2022-cloze} & 2,335 & Manually built and validated by native  & -- & -- & -- & --\\
      XCOPA~\cite{ponti-etal-2020-xcopa} &  600 & Translated from English data  & -- & -- & -- & --\\
      XStoryCloze~\cite{lin-etal-2022-shot} & 1,872 & Translated from English data  & -- & -- & -- & --\\
      \bottomrule
    \end{tabular}
    }
    \caption{Comparison of \datasetname{} with other cultural knowledge and reasoning datasets containing instances in Indonesian. The metadata includes \textbf{Size} (number of Indonesian instances), \textbf{Cultural?} (whether the data considers cultural nuances), \textbf{Location?} (whether the data includes fine-grained location information, such as provinces, as context), \textbf{\#province} (number of Indonesian provinces covered), and \textbf{\#topic} (number of fine-grained topics covered). * indicates the dataset involves question generation with less emphasis on reasoning.}
    \label{tab:dataset_comparison}
\end{table*}

\textit{Can large language models effectively reason based on the diverse cultures of Indonesia?} To capture the rich diversity of Indonesian cultures, we predefined 12 fine-grained topics as guidelines for data construction. Figure~\ref{fig:topic} displays the topic distribution in \datasetname{}, with the majority focusing on food, weddings, art, pregnancy and children, and family relationships. Additionally, we also pose the question: \textit{Is there any influence of geographical location on the commonsense reasoning of language models?} We address these questions through comprehensive experiments across different language models, incorporating several levels of location granularity as additional context in the prompt.

Our contributions can be summarized as follows:
\begin{itemize}
    \item We present \datasetname{}, a high-quality cultural reasoning dataset in the Indonesian language, covering eleven provinces of Indonesia and twelve fine-grained cultural topics. Our dataset has 2,429 instances, and was developed by local people with rigorous quality controls in place.
    \item We assess 19 open-weight multilingual models, 6 open-weight Indonesian-centric models, and 2 closed-weight models. Although local individuals can answer all questions correctly (i.e., 100\% accuracy), most open-weight models struggle to comprehend Indonesian cultures. Interestingly, we observed that Llama--3~\cite{dubey2024llama} 
    is competitive with GPT--4 \cite{OpenAI2023GPT4TR}.
    \item We conduct a thorough analysis over various dimensions: (1) model performance for each province and topic; (2) the influence of different granularities of location context (i.e., none, province, country); (3) model performance over English translations; and (4) analysis of model explanations for a given answer.
\end{itemize}

\section{Related Work}

\paragraph{Commonsense Reasoning in English} Many studies have focused on commonsense reasoning in English, often overlooking considerations of culture and geographical location.  Early work included the \texttt{Winograd Schema Challenge} \cite{levesque2012winograd} and \texttt{WinoGrande} \cite{sakaguchi2021winogrande} for pronoun coreference resolution. Other research areas include reasoning based on cause-effect relationships \cite{roemmele2011choice}, physical activities \cite{DBLP:conf/aaai/BiskZLGC20}, social interactions \cite{sap-etal-2019-social}, cloze story completion \cite{mostafazadeh-etal-2016-corpus}, sentence completion \cite{zellers-etal-2019-hellaswag}, numerical reasoning \cite{lin-etal-2020-birds}, and temporal reasoning \cite{qin-etal-2021-timedial}. Additionally, pretrained language models have been employed in other work to extract structured commonsense knowledge by providing seed words \cite{davison-etal-2019-commonsense}, and using \textit{code} language models \cite{madaan-etal-2022-language}.


\paragraph{Cultural Commonsense Reasoning with Geographical Contexts}  Previous studies have explored commonsense reasoning with geographical context. \citet{shwartz-2022-good} investigated time perception (e.g., morning and night) across different locations, while \citet{yin-etal-2022-geomlama} examined cultural knowledge of language models across five countries using datasets built from templates and translations. Other work has focused on automatically extracting cultural knowledge from various sources, including Wikipedia \cite{fung2024massively}, conversations \cite{fung-etal-2023-normsage}, and Common Crawl \cite{nguyen2023extracting}, incorporating location context with the assistance of large language models. Related, \citet{ziems-etal-2023-normbank} created a knowledge bank for situational norms, using English-speaking Mechanical Turk annotators and incorporating a country taxonomy. Unlike this work, \datasetname{} specifically concentrates on cultural reasoning across Indonesian provinces, developed and validated manually by local people (experts). Compared to the automatic method and English-speaking crowd workers for data construction, \datasetname{} arguably contains less noise, and is free from the training data contamination of large language models (LLMs).





\paragraph{Commonsense Reasoning with Indonesian contexts}

Table~\ref{tab:dataset_comparison} shows a comparison of \datasetname{} with other Indonesian datasets for cultural knowledge and reasoning evaluation. Commonsense reasoning in Indonesian language models has been studied using translated English--Indonesian datasets, such as \texttt{XCOPA} \cite{ponti-etal-2020-xcopa} and \texttt{XStoryCloze} \cite{lin-etal-2022-shot}. However, these datasets potentially introduce a cultural bias toward culture grounded in the English language. \texttt{IndoCloze} \cite{koto-etal-2022-cloze} was the first commonsense reasoning dataset in Indonesian, developed by native Indonesian workers following the cloze story completion framework \cite{mostafazadeh-etal-2016-corpus}. However, \texttt{IndoCloze} lacks local cultural nuances and fine-grained geographical context. \citet{wibowo-etal-2024-copal} followed the COPA framework \cite{roemmele2011choice} to build a dataset with contexts limited to Jakarta. In other work, \citet{afina2024can} studied the capability of LLMs in generating questions with cultural norms, for both general Indonesian and specific Sundanese contexts, while \citet{liu-etal-2024-multilingual} used proverbs and LLMs to generate conversational data. In contemporary work, \citet{myung2024blend} released BLEnD, a large-scale cultural knowledge dataset, built using templates, translation, and human validations, covering the West Java province in Indonesia. BLEnD specifically focuses on short-answer questions, limiting its capacity for reasoning evaluation. 
Unlike most other datasets that do not consider geographical factors, \datasetname{} has broad coverage across eleven provinces, thereby providing greater inclusivity for local communities in Indonesia.



\section{{IndoCulture}}
\label{sec:data_construct}

As illustrated in Figure~\ref{fig:map}, \datasetname{} is a sentence completion task in the Indonesian language featuring a one-sentence premise, three plausible options, and one correct option to evaluate reasoning ability and cultural knowledge across eleven Indonesian provinces.
While sentence completion tasks are straightforward for humans, answering \datasetname{} requires machines to engage in cultural reasoning to logically conclude which of the three options is logically consistent with the first sentence \cite{huang-chang-2023-towards}. The dataset includes a total of 2,429 instances.

\subsection{Data Construction}
\datasetname{} was constructed manually by humans, and verified through a two-step process.

\paragraph{Worker Recruitment}

Culture generally arises from the shared experiences, traditions, and beliefs of a specific group over time, often closely intertwined with native populations. With this in mind, we engaged individuals from various provinces across Indonesia to assist in preparing data for the \datasetname{} benchmark. 

During recruitment, we presented a few examples of the intended \datasetname{} data and requested each candidate to generate similar instances tailored to the context of their respective provinces. From a pool of 58 applicants, we carefully selected 22 expert workers representing 11 provinces (with 2 workers selected per province). These recruited expert workers are local residents and have resided in their respective provinces for a minimum of 10 years, thereby possessing a profound understanding of local customs and culture. The age range of our workforce spans from 21 to 35 years old, with educational backgrounds distributed as follows: 3 high school graduates, 14 bachelor's degree holders, 4 Master's degree holders, and 1 PhD holder. 

During data construction, each expert worker fulfilled the dual roles of instance writer and quality controller. Each worker was compensated above the monthly minimum wage in Indonesia.


\paragraph{Province Selection}
The provinces covered in this study represent the diversity of Indonesian cultures. The 11 provinces (in Figure~\ref{fig:map}) are spread across 6 islands of the Indonesian archipelago,  which are inhabited by different ethnic groups who speak different regional languages and adhere to different religions. 

\paragraph{Topic Taxonomy}
\datasetname{} consists of 12 topics and 66 fine-grained subtopics, carefully constructed based on discussions and brainstorming with Indonesian natives. The selection of these topics and subtopics was guided based on several criteria and motivations: (1) relevance to Indonesian culture; (2) diversity and coverage; (3) regional representation (e.g., religious holidays); (4) practicality; and (5) expert consultation (i.e., native speaker feedback). Compared to the other Indonesian datasets in Table~\ref{tab:dataset_comparison}, \datasetname{} includes a richer array of fine-grained topics. Below is a list of the topics along with their detailed subtopics. The numbers following each topic indicate the total number of instances required to be written by one worker (with a total of 150 per worker). 

\begin{enumerate}
    \item \textbf{Food} (22): breakfast (2); lunch (3); dinner (2); snacks (2); food souvenirs (3); traditional foods and beverages (5); eating habits (1); cutlery (1); cooking ware (1),  
    fruit (2).
    \item \textbf{Wedding} (20): traditions before marriage (3); traditions when getting married (3); traditions after marriage (3); men's wedding clothes (2); women's wedding clothes (2); invited guests (2); wedding location (1); foods at a wedding (2); gifts brought to weddings (2).
    \item \textbf{Family relationship} (13): relationships within the main family (3); relationships in the extended family (3); relations with society/neighbors (5); 
    clan/descendant system (2).
    \item \textbf{Pregnancy and kids} (16): traditions during pregnancy (4); traditions after birth (2); how to care for a newborn baby (2); how to care for toddlers (2); how to care for children (2); how to care for teenagers (2); parents and children interactions as adults (2).
    \item \textbf{Death} (10): when death occurs (2); the process of dealing with a corpse (2); traditions after the body is buried (2); the clothes of the mourners (2); inheritance matters (2).
    \item \textbf{Religious holiday} (12): traditions before religious holidays (2); traditions leading up to religious holidays (4); traditions during religious holidays (5); traditions after religious holidays (1).   
    \item \textbf{Agriculture} (6): what to plant (2); traditions when planting (2); harvest (2).
    \item \textbf{Fisheries and trade} (7): traditions of taking care of livestock/fish (5); buying and selling traditions (2)
    \item \textbf{Art} (16): musical instruments (3); folk songs (3); traditional dances (3); use of art at certain events (5); poetry or similar literature (2)
    \item \textbf{Traditional games} (5): game types; (3), location played (2).
    \item \textbf{Daily activities} (10): morning activities (1); afternoon activities (1); evening activities (1); leisure activities (3); house, household, and transportation (4).
    \item \textbf{Socio-religious aspects of life} (13): regular religious activities (2); mystical things (2); traditional ceremonies (1); lifestyle (3); self care (1); traditional medicine (3); traditional sayings (1).
\end{enumerate}

\paragraph{Instance Writing}
For each instance, workers were asked to craft two culturally-relevant sentences that align with the predefined subtopic. The first sentence serves as the premise context, while the last sentence acts as the correct answer. Subsequently, the annotator generates two additional plausible sentences as distractors by modifying cultural objects or activities from the correct sentence. These distractors are designed to reflect local cultural contexts, ensuring they are challenging yet unambiguous, and could potentially serve as correct answers in other regional contexts. Workers were given a period of two months to complete the task.
\paragraph{Two Stages of Quality Control}


In stage 1, we implemented quality control by pairing two annotators from the same province. Each annotator was tasked with answering a set of questions prepared by the other annotator, and vice versa. During this phase, the annotator were presented with a premise sentence and three shuffled options. They were allowed to search for the answer from any source if they were unsure. Instances that were incorrectly answered by the second annotator were discarded, as we hypothesize that these instances may contain incorrect answers or possess a level of ambiguity. Additionally, annotators were required to identify whether the instance is province-specific (binary annotation: True/False), indicating that it is uniquely relevant in their province and not in others. 


In stage 2 of quality control, the first two authors of this paper performed post-editing of data that passed the first stage of quality control. We first focused on correcting the linguistic aspects of the text, including checking for spelling errors. Although the text is written in Indonesian, some annotators may use dialects or be influenced by the structure or style of regional languages. In these cases, we corrected the text to adhere to Indonesian grammar.

To maintain the quality of \datasetname{}, we rigorously filtered instances that contained: (1) poor writing, in the case that it was difficult to post-edit to enhance their quality; (2) obvious answer options, which allow for easy guessing of the correct choice without understanding the cultural context; and (3) ambiguous contexts, where all options are equally valid as the correct answer. For example, in a topic about breakfast, the three options might include one traditional food alongside two other very commonly consumed foods in Indonesia, and be considered too obvious.

Furthermore, we manually verified the province-specific annotations for each instance using the Google search engine. We annotated whether the instance pertains to national-level culture or not. If the example is specific to a province, it will be annotated as uncommon in national culture, and vice versa. 





\subsection{Data Statistics}

After the instance writing process, we initially collected 3,162 instances out of a target of 3,300 instances (22 workers $\times$ 150 subtopics). Although we requested each annotator to produce 150 instances, not all were able to complete their allotted tasks within the given timeframe. Unfortunately, we were unable to find additional candidates from the same local province to address the data deficiencies~\cite{winata-etal-2023-nusax}. 

\begin{figure}[t]
    \centering
    \includegraphics[width=\linewidth]{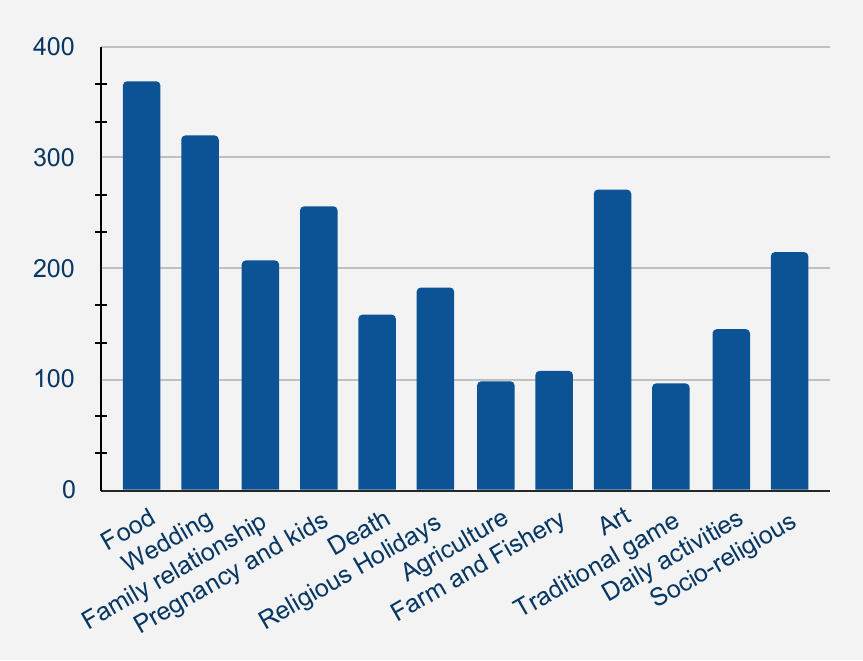} 
    \caption{Topic distribution in \datasetname{}.}
    \label{fig:topic}
    \vspace{0.3cm}
\end{figure}

In stage 1 of quality control, the initial pool of 3,162 instances was reduced to 2,801 instances, and stage 2 of quality control further reduced the sample to 2,429 high-quality samples. The data distribution of \datasetname{} per province is presented in Table~\ref{tab:stat}. Approximately three-quarters of \datasetname{} instances contain province-specific content, with an average length of around 35 words. \datasetname{} covers multiple topics, as illustrated in Figure~\ref{fig:topic}.

\begin{table}[t]
    \centering
    \resizebox{\linewidth}{!}{
        \begin{tabular}{lcccc}
        \toprule
        \multirow{2}{*}{\textbf{Province}} & \multirow{2}{*}{\textbf{\#}} & \textbf{province} &  \multirow{2}{*}{\textbf{$\mu$(word)}} & \multirow{2}{*}{\textbf{$\mu$(char)}} \\
        & & \textbf{specific (\%)} & & \\
        \midrule
        Aceh & 246 & 70.7 & 28.0 & 175.9 \\
        North Sumatra & 234 & 83.8 & 36.8 & 246.0 \\
        West Sumatra & 299 & 74.6 & 39.6 & 261.4 \\
        West Java & 231 & 58.0 & 37.5 & 244.8 \\
        Central Java & 171 & 66.7 & 39.3 & 260.5 \\
        East Java & 233 & 69.5 & 46.0 & 310.4 \\
        Bali & 241 & 76.3 & 33.3 & 216.1 \\
        NTT & 103 & 72.8 & 31.8 & 203.6 \\
        South Borneo & 233 & 83.7 & 33.3 & 226.0 \\
        South Sulawesi & 185 & 90.3 & 33.6 & 227.8 \\
        Papua & 253 & 88.1 & 37.3 & 245.0 \\
        \midrule
        All & 2429 & 76.0 & NA & NA \\
        \bottomrule
        \end{tabular}
    }
    \caption{Overall statistics of \datasetname{} by province.}
    \vspace{0.cm}
    \label{tab:stat}
\end{table}

\section{{Experiments}}

\subsection{Set-Up}

We evaluate 27 language models in zero-shot settings: (1) nineteen open-weight multilingual language models of varying sizes, namely BLOOMZ \cite{muennighoff-etal-2023-crosslingual}, mT0 \cite{muennighoff-etal-2023-crosslingual}, Bactrian-X \cite{li2023bactrian}, Llama--2 \cite{touvron2023llama}, and Llama--3~\cite{dubey2024llama}; (2) two South East Asian language models, namely SeaLLM \cite{nguyen-etal-2024-seallms}, and SeaLion \cite{sea_lion_2023}; (3) four Indonesian-centric language models, namely IndoBART \cite{cahyawijaya2021indonlg}, IndoGPT \cite{cahyawijaya2021indonlg}, Merak \cite{Merak}, and Komodo \cite{komodo}; (3) two closed-weight models, namely  ChatGPT: \texttt{gpt-3.5-turbo} \cite{ouyang2022training} and GPT--4: \texttt{gpt-4-0613} \cite{OpenAI2023GPT4TR}. Please refer to Appendix~\ref{ap:model} for further details.

\begin{figure}[t]
    \centering
    \includegraphics[width=0.8\linewidth]{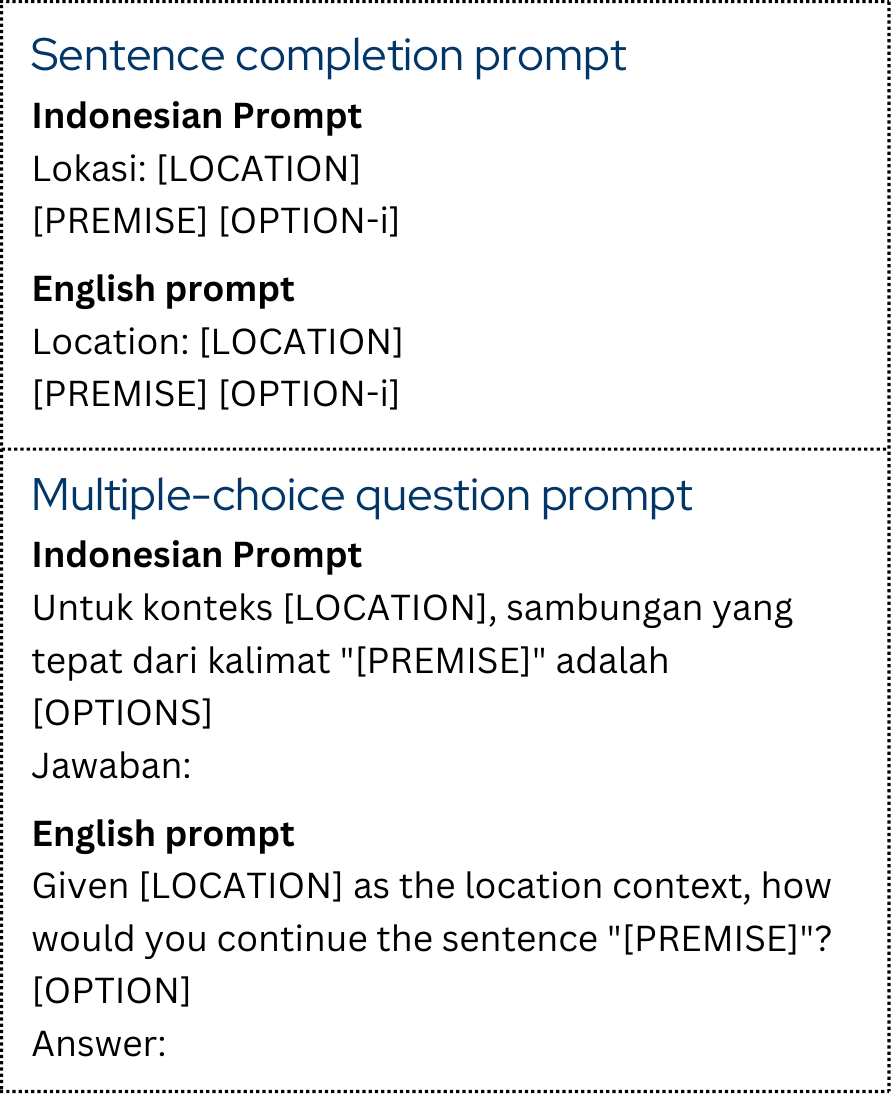} 
    \caption{Templates for sentence completion and multiple-choice questions prompts.}
    \label{fig:prompt}
    \vspace{-0cm}
\end{figure}

First, we evaluate the effectiveness of sentence completion and multiple-choice question strategies in predicting the correct options using the Indonesian and English prompt templates shown in Figure~\ref{fig:prompt}. In both scenarios, we conduct benchmarks across three distinct location contexts. Formally, given a premise $s$, three candidate options $c_1$, $c_2$, $c_3$, and location $l\in\{\texttt{none}, \texttt{Indonesia}, province\}$, for sentence completion, we select the correct option based on:
\begin{align*}
\underset{c}{\mathrm{argmax}} \log P(\text{concat}(s, c) | l)
\end{align*}
Here, $\text{concat}(s,c)$ denotes the concatenation of premise $s$ and candidate option $c$, separated by a space. In the case of multiple-choice questions, we devise a template for the prompt question and determine the answer by selecting the option with the highest probability among letters A, B, and C. 

For GPT--3.5 and GPT--4, we exclude experiments with sentence completion because the closed-weight models do not provide an overall probability score. For multiple-choice questions, we select the first generated token that corresponds to the letters A, B, or C using a regular expression.

\subsection{Results}

\begin{table*}[t]
    \centering
    \resizebox{0.8\linewidth}{!}{
        \begin{tabular}{lcccccccc}
        \toprule
            \multirow{2}{*}{\textbf{Model (\#parameter)}} & \multicolumn{3}{c}{\textbf{Completion}} & & \multicolumn{3}{c}{\textbf{MCQ}} \\
            \cmidrule{2-4} \cmidrule{6-8}
             & $l=\text{None}$ & $l=\text{Ind}$ &$l=\text{Prov}$ &  & $l=\text{None}$ & $l=\text{Ind}$ &$l=\text{Prov}$ \\
             \midrule
            Human &  -- &  -- & 100 &  &  -- &  -- &  100.0 \\
            Random & 33.3 & 33.3 & 33.3 &  & 33.3 & 33.3 &  33.3 \\
            \midrule 
            BLOOMZ (560M) & 37.2 & 35.3 & 35.3 &  & 32.5 & 32.4 & 32.5\\
            BLOOMZ (1.1B) & 36.3 & 36.9 & 37.2 &  & 32.4 & 32.4 & 32.4\\
            BLOOMZ (3B) & 38.6 & 40.7 & 41.5 &  & 47.0 & 48.6 & 49.2\\
            BLOOMZ (7B) & 41.3 & 44.1 & 44.6 &  & 49.5 & 50.6 & 50.5 \\
            \hdashline
            mT0$_{\text{small}}$ (300M) & 28.3 & 28.1 & 28.3 &  & 34.1 & 33.1 & 32.6 \\
            mT0$_{\text{base}}$ (580M) & 28.4 & 28.1 & 28.5 &  & 35.4 & 35.1 & 35.6 \\
            mT0$_{\text{large}}$ (1.2B) & 29.6 & 29.5 & 30.1 &  & 35.6 & 35.7 & 35.8 \\
            mT0$_{\text{xl}}$ (3.7B) & 31.9 & 31.0 & 31.2 &  & 49.8 & 50.5 & 50.7 \\
            mT0$_{\text{xxl}}$ (13B) & 33.2 & 33.5 & 34.3 &  & 52.7 & 51.4 & 52.1 \\
            \hdashline
            Bactrian-X$_{\text{LLaMA}}$ (7B) & 33.8 & 34.2 & 34.2 &  & 38.0 & 38.6 & 38.9 \\
            Bactrian-X$_{\text{LLaMA}}$ (13B) & 33.3 & 35.2 & 35.1 &  & 38.6 & 38.2 & 38.6 \\
            \hdashline
            Llama--2 (7B) & 37.2 & 37.5 & 37.7 &  & 40.5 & 39.9 & 38.8 \\
            Llama--2 chat (7B) & 37.3 & 37.4 & 37.9 &  & 40.6 & 41.3 & 40.7 \\
            Llama--2 (13B) & 39.6 & 40.2 & 40.2 &  & 47.6 & 47.6 & 47.3 \\
            Llama--2 chat (13B) & 38.6 & 38.9 & 39.3 &  & 47.8 & 49.6 & 49.6 \\
            \hdashline
            Llama--3 (8B) & 41.0 & 42.2 & 43.4 &  & 54.4 & 54.4 & 55.1 \\
            Llama--3 Instruct (8B) & 41.9 & 41.5 & 42.3 &  & 56.7 & 57.6 & 59.0 \\
            Llama--3 (70B) & \textbf{51.2} & \textbf{51.7} & \textbf{54.3} &  & \textbf{68.6} & \textbf{69.9} & 72.7 \\
            Llama--3 Instruct (70B) & 49.2 & 49.6 & 52.2 &  & 68.5 & 69.3 & \textbf{73.3} \\
            \midrule 
            IndoBART (132M) & 42.4 & 41.3 & 42.1 &  & 32.6 & 32.4 & 32.7\\
            IndoGPT (117M) & 42.6 & 41.9 & 42.4 &  & 33.7 & 33.8 & 34.7\\
            Merak (7B) & 41.0 & 41.5 & 43.5 &  & 51.9 & \textbf{53.1} & \textbf{53.2}\\
            SeaLLM (7B) & 39.1 & 39.3 & 41.1 &  & \textbf{52.2} & \textbf{53.1} & 53.0\\
            SEA-LION (7B) & 38.8 & 38.9 & 39.7 &  & 33.8 & 33.0 & 33.3\\
            Komodo (7B) & \textbf{45.1} & \textbf{45.4} & \textbf{46.1} &  & 37.6 & 35.1 & 36.1\\
            \midrule 
            GPT--3.5 (NA) &  -- &  -- &  -- &  & 59.8 & 60.9 & 62.7\\
            GPT--4 (NA) &  -- &  -- &  -- &  & \textbf{69.1} & \textbf{71.8} & \textbf{75.9}\\
    
            \bottomrule
        \end{tabular}
    }
    \caption{Zero-shot accuracy across various models and settings. ``MCQ'' refers to the multiple-choice question method, and $l$ denotes the location as additional context (``Ind'' and ``Prov'' denote the country of Indonesia, and the corresponding province). The bold numbers highlight the highest score within each model group.}
    \label{tab:result}
\end{table*}

\paragraph{Overall observation} The results presented in Table~\ref{tab:result} display the performance across various models and settings. The overall observation is that most open-weight models struggle  to understand Indonesian culture, contrasting sharply with the 100\% accuracy achieved by humans (i.e., natives of the given province). Among open-weight models, Llama--3 achieves the highest accuracy of 73.3\%. Other open-weight models such as Merak and mT0$_\texttt{xxl}$ achieve accuracy of 52--53\%, while closed-weight models, such as GPT--3.5 and GPT--4, achieve performances of 62.7\% and 75.9\%, respectively. These findings underscore the challenging nature of the \datasetname{} dataset.

\paragraph{The multiple-choice question method is generally better.} Our findings suggest that the multiple-choice question method tends to outperform the sentence completion method, with exceptions noted for BLOOMZ (560M, 1.1B), IndoBART, IndoGPT, and Komodo. Interestingly, in the sentence completion task, the Indonesian-focused language model Komodo (7B) outperforms nearly all large multilingual models, with the exception of Llama--3 (70B). However, Komodo experiences a significant decline for the multiple-choice question method, with a notable margin of 10--12 points. This discrepancy could potentially be attributed to differences in the nature of language model training and instruction-tuning.

\paragraph{Impact of location context on model performance} Our investigation reveals that incorporating various levels of location granularity has a noticeable effect on zero-shot performance, especially for models with larger parameter sizes. Detailed location context notably enhances the accuracy of BLOOMZ (7B), Llama--2 (13B), Llama--3 (70B), Merak (7B), SeaLLM (7B), Komodo (7B), GPT--3.5, and GPT--4. For instance, in GPT--4, the accuracy gap between $l=\texttt{none}$ and $l=\texttt{Indonesia}$ is 2.7, and this gap further increases by 7 points when $l=\texttt{Province}$ is assigned.

\subsection{Analysis}
\label{sec:analysis}

Given the exceptional performance of models with large parameter sizes using the multiple-choice question method and location $l=\texttt{Province}$, we employ these configurations for our analysis. In this section, our main focus is on the top three performing models: Llama--3 Instruct (70B), Merak (7B), and GPT--4. These models represent a multilingual open-weight model, Indonesian-centric open-weight model, and closed-weight model, respectively.

\begin{table}[t]
    \centering
    \resizebox{\linewidth}{!}{
        \begin{tabular}{lccccccc}
        \toprule
        \multirow{2}{*}{{\textbf{Province}}} & \multicolumn{2}{c}{\textbf{Merak}} & \multicolumn{2}{c}{\textbf{Llama--3}} &  \multicolumn{2}{c}{\textbf{GPT--4}}  \\
        \cmidrule{2-7}
         & \textbf{$\neg$PS }& \textbf{PS} & \textbf{$\neg$PS} & \textbf{PS} & \textbf{$\neg$PS} & \textbf{PS} \\
        \midrule
        Aceh              & 59.7 & 53.4 & {77.8} & {68.4} & \ok{93.1} & \ok{73.6} \\
        North Sumatra     & \no 58.3 & 46.9 &  \no 69.4 & 67.5 & \no 75.0 & 73.2 \\
        West Sumatra      & \no 50.0 & \no 41.3 & \no 73.7 &  \no 64.6 & 85.5 & \no 63.7 \\
        West Java         & 65.3 & \ok{58.8} & \ok{91.6} & \ok{86.3} & \ok{92.6} & \ok{81.7} \\
        Cental Java       & 58.8 & 47.1 & 79.4 &  67.1  & 82.4 & 72.9 \\
        East Java         & 57.4 & \no 37.6 &  88.9 & \no 57.4 & 87.0 & \no 63.4 \\
        Bali              & \ok{78.9} & \ok{65.6} & \ok{96.5} & \ok{86.7} & \ok{93.0} & \ok{89.4} \\
        NTT               & 64.3 & 52.0 &  78.6 &  \no 61.3 & 85.7 & 68.0 \\
        South Borneo      & \ok{65.7} & \no 46.4 & 85.7 & 66.9 & \no 77.1 & \no 67.4 \\
        South Sulawesi    & \no 53.3 & 47.9 & \no 73.3 & 66.9 & \no 80.0 & 70.6 \\
        Papua             & \ok{76.7} & \ok 55.2 & \ok 90.0 & \ok 73.1 & 90.0 & 71.3 \\
        \bottomrule
        \end{tabular}
    }
    \caption{Top-3 model accuracy by province. ``PS'' indicates instances containing province-specific context, while ``$\neg$PS'' indicates otherwise. The green and red cells indicate the top three and bottom three scores, respectively.}
    \label{tab:result_province}
\end{table}

\paragraph{Results by province} Table~\ref{tab:result_province} highlights that the top 3 performing LLMs exhibit a nuanced understanding of culture within Indonesian provinces, particularly excelling in the cultures of West Java and Bali compared to other provinces. Llama--3 and GPT--4, for instance, achieve the best accuracy of more than 90\%, while Merak achieves the best performance in Bali, Papua, and West Java, with accuracies ranging between 55\% and 79\%. In other provinces like West Sumatra and South Borneo, the models typically exhibit poorer performance. Specifically, for Llama--3, the performance gap compared to Bali ranges from 10 to 30\%. This highlights the presence of cultural biases and a lack of inclusivity in model reasoning abilities, likely stemming from the distribution of training data. The proximity of West Java to Jakarta (Indonesia's capital) and Bali's global status as a tourism destination may contribute to the abundance of textual data on these two cultures.

We also note a consistent disparity between non-province and province-specific contexts across all models, with models generally finding non-province contexts easier to comprehend. On average, this gap ranges from 12 to 13 points for the three models, highlighting the challenge posed by province-specific content and emphasizing the significant influence of location context on the reasoning ability of LLMs.

\paragraph{Results by topic} Table~\ref{tab:result_topic} shows the accuracy of the top 3 performing models across different topics. Similar to Table~\ref{tab:result_province}, the models perform worse in province-specific contexts for all topics, with the notable exception of food. For province-specific contexts, GPT--4 excels for the themes of food, religious holidays, and arts while for non-specific contexts, Llama--3 achieves accuracy more than 90\% for agriculture, daily activities, and religious holidays.

\begin{table}[t]
    \centering
    \resizebox{\linewidth}{!}{
        \begin{tabular}{lccccccc}
        \toprule
        \multirow{2}{*}{{\textbf{Topic}}} & \multicolumn{2}{c}{\textbf{Merak}} & \multicolumn{2}{c}{\textbf{Llama--3}} &  \multicolumn{2}{c}{\textbf{GPT--4}}  \\
        \cmidrule{2-7}
         & \textbf{$\neg$PS }& \textbf{PS} & \textbf{$\neg$PS} & \textbf{PS} & \textbf{$\neg$PS} & \textbf{PS} \\
        \midrule
        Food & 58.7 & 50.2 & \no 69.6 & 71.8 & \no 73.3 & \ok 77.3 \\
        Wedding & 60.0 & 50.2 & 80.0 & \no 66.9 & 82.5 & 66.9 \\
        Family relationship & \no 55.1 & \no 39.6 & \no 75.5 & \no 66.9 & 85.2 & \no 63.0 \\
        Pregnancy and kids & 57.1 & \no 44.3 & 83.5 & 69.0 & 84.7 & \no 64.6 \\
        Death & \no 56.9 & 51.0 & 86.2 & \no 62.5 & \no 79.4 & \no 63.5 \\
        Religious holidays & 66.7 & \ok 64.3 & 79.7 & 78.6 & \ok 91.5 & \ok 78.6 \\
        Agriculture & 65.5 & \ok 54.5 & \ok 100.0 & \ok 80.3 & \ok 93.8 & 72.7 \\
        Farm and fishery & 67.9 & \no 42.9 & 85.7 & \no 62.3 & 80.0 & 72.7 \\
        Art & \ok 77.8 & \ok 55.2 & 83.3 & \ok 72.2 & 90.0 & \ok 82.5 \\
        Traditional game & \no 38.5 & 48.8 & \no 69.2 & 68.3 & \no 78.6 & \no 64.6 \\
        Daily activities & \ok 68.5 & 52.3 & \ok 92.6 & \ok 72.7 & \ok 93.1 & 76.1 \\
        Socio-religious & \ok 70.5 & 51.0 & \ok 88.5 & 71.8 & 86.4 & 75.8 \\
        \bottomrule
        \end{tabular}
    }
    \caption{Top-3 model accuracy by topic. ``PS'' indicates instances containing province-specific contexts, while ``$\neg$PS'' indicates otherwise. The green and red cells indicate the top three and bottom three scores, respectively.}
    \label{tab:result_topic}
\end{table}

\paragraph{Results by fine-grained cultural elements} We tasked two expert workers with annotating 200 random samples from \datasetname{} based on six cultural elements, as derived from \citet{axtell1998gestures,williams2014keywords}.\footnote{\url{https://www.languageeducatorsassemble.com/elements-of-culture}}\footnote{\url{https://pressbooks.howardcc.edu/soci101/chapter/3-2-the-elements-of-culture/}} While these elements may not encompass every cultural aspect, we contend that they cover the most prominent or pivotal elements, including: (1) \textit{symbols} (material or non-material objects representing meaning); (2) \textit{artifacts} (material or non-material objects produced by society); (3) \textit{values and beliefs} (principles, ideas, and concepts assumed to be ideal and correct in society); (4) \textit{norms} (rules guiding values and beliefs); (5) \textit{language}; (6) \textit{rituals} (established procedures and ceremonies); and (7) \textit{other}, for examples that do not fit into any of the defined elements. This annotation is a multi-label task, and the average Kappa score across the cultural elements is 0.56, with each ranging from 0.4 to 0.75. These scores indicate moderate to substantial agreement.

Table~\ref{tab:culture_element} displays the distribution of each cultural element in our dataset, along with the performance breakdown across Merak, Llama--3, and GPT--4. Among the 200 random samples, we observe that 42.5\% of our data contains \textit{artifacts}, 37.5\% \textit{norms}, and 30\% \textit{rituals}. Only 4\% of the data pertains to \textit{symbols}, while 7.5\% belongs to the \textit{other} category. Merak shows lower accuracy in \textit{norms}, with a 24\% decrease compared to \textit{values and beliefs}. Conversely, Llama--3 performs best in \textit{values and beliefs} with 85\% accuracy, but accuracy drops by 23\% for \textit{norms}. GPT--4 maintains relatively stable accuracies across cultural elements, with differences averaging between 3\% and 5\%. Furthermore, \textit{language} presents a challenge for Merak, achieving only 38\% accuracy, whereas Llama--3 and GPT--4 achieve 66\% and 72\% accuracy, respectively.

\begin{table}[t]
    \centering
    \resizebox{\linewidth}{!}{
        \begin{tabular}{L{4cm}C{1cm}C{1.5cm}C{1.2cm}}
        \toprule
        \textbf{Cultural element (\%)} & \textbf{Merak} & \textbf{Llama--3} & \textbf{GPT--4} \\
        \midrule
        Symbols (4) & 50.0 & 50.0 & 70.8 \\
        Artifacts (42.5) & 55.3 & 74.1 & 67.8 \\
        Values and Beliefs (10.5) & 61.9 & 85.7 & 69.8 \\
        Norms (37.5) & 38.7 & 62.7 & 73.6 \\
        Language (19.5) & 38.5 & 66.7 &  72.0 \\
        Ritual (30) & 53.3 & 65.0 &  70.7 \\
        Other (7.5) & 66.7 & 66.7 & 69.2 \\
        \bottomrule
        \end{tabular}
    }
    \caption{Accuracy comparison of Merak, Llama--3--Instruct (70B), and GPT--4 across 200 random samples, categorized by cultural elements. The numerical value following each cultural element indicates its proportion within the samples.}
    \label{tab:culture_element}
    \vspace{0.5cm}
\end{table}

\paragraph{Can the model provide a reasonable explanation to support the answer?}

\begin{figure}[t]
    \centering
    \includegraphics[width=\linewidth]{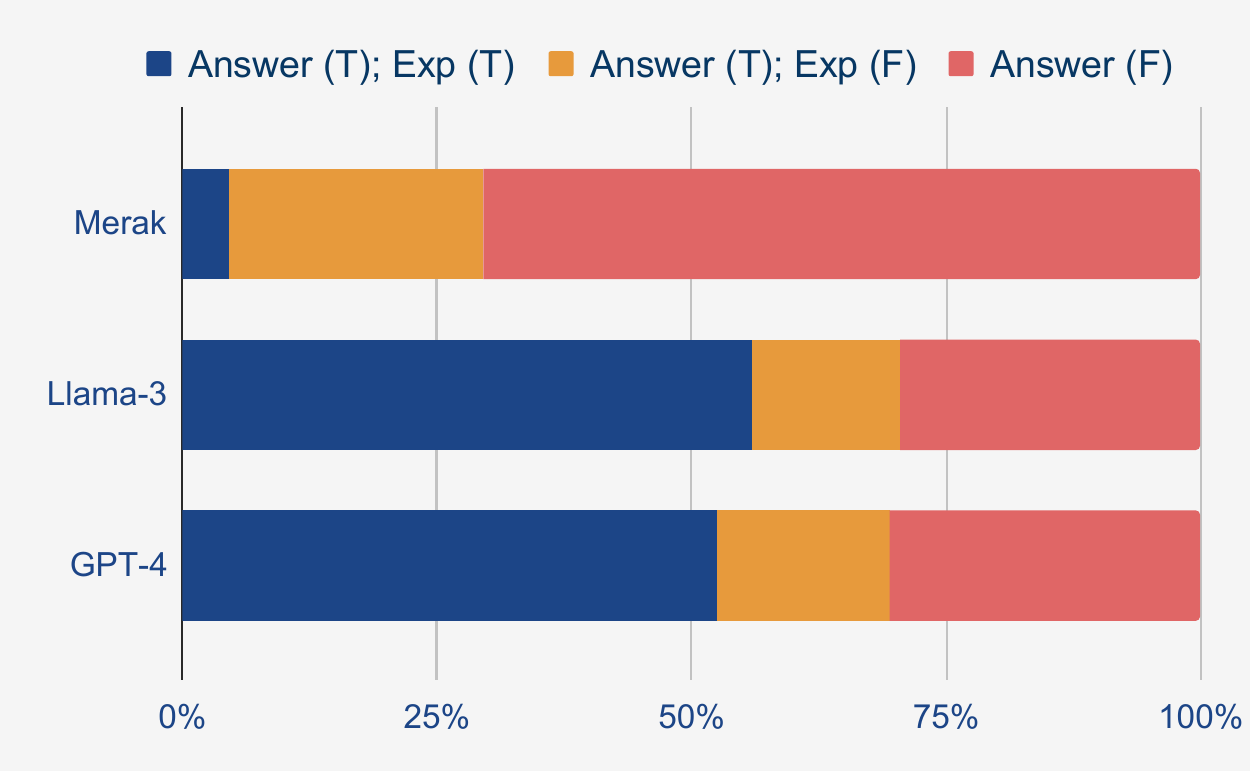} 
    \caption{Performance comparison between Merak (7B), Llama--3 Instruct (70B), and GPT--4 based on text generation output. ``\texttt{Answer (T)}'' indicates that the generated answer is true, while ``\texttt{Exp(F)}'' denotes that the answer explanation is false.}
    \label{fig:ana}
\end{figure}

We conduct a manual investigation of the text generation output for Merak (7B), Llama--3, and GPT--4 across 200 random samples. This involves manually examining the generated answer along with its explanation. To obtain the explanation, we modify the Indonesian prompt in Figure~\ref{fig:prompt} by adding the string \textit{Jelaskan jawabanmu!} ``Explain your answer!''. Our annotation process is binary, categorizing explanations as either True or False. We label an explanation as False if it is absent, contains hallucinations, or provides inaccurate information.\footnote{We use the Google search engine to verify the correctness of the explanation.}

As anticipated, there is a substantial drop in accuracy for Merak (7B) from 53.2\% (as shown in Table~\ref{tab:result}) to 29.5\%. This discrepancy underscores the limitations of relying solely on token probabilities to assess the true capability of a language model. Interestingly, only 4.5\% of the samples are answered correctly with the appropriate explanation by Merak, despite it being the top performer among the Indonesian-centric language models. Larger models like Llama--3 and GPT--4 achieve more robust accuracies of 70.5\% and 69.5\%, which are 3--5\% lower than those indicated in Table~\ref{tab:result}. However, both models encounter challenges in generating appropriate explanations for correctly-answered samples, with Llama--3 and GPT--4 producing explanation errors in 29\% and 34\% of cases, respectively.

\paragraph{Does language affect model performance?} We automatically translated \datasetname{} to English using the Google Translate API\footnote{Accessed in March 2024.} and used the English prompt in Figure~\ref{fig:prompt} to evaluate the models. Specifically for this part, we include more models for comparison. All results over English text dropped, except for Llama--2 and Merak. This could be attributed to two reasons. First, Llama--2 is an English-centric model, and Merak is fine-tuned from Llama--2. Second, the performance drop for other models could be caused by translation errors. We further investigated this with 100 random samples and found that 81 samples had acceptable translations. We observed translation errors such as pronoun mismatches, inaccurate proverb translations, and inaccurate translations of local terms, such as \textit{pupuik} translated as ``fertilizer''.\footnote{\textit{pupuik} is a traditional musical instrument in West Sumatra. It is worth noting that the word \textit{pupuik} closely resembles the word \textit{pupuk} in Indonesian, which means fertilizer.} To better understand the cultural gap in language models, we followed the approach of \citet{liu-etal-2024-multilingual} to manually correct the translations and reevaluate the models. We found that GPT--4's performance over the 100 random samples was 77.0 for the original Indonesian text, 68.0 for the English machine-translated text, and 72.0 for the English translation fixed by humans.

\begin{figure}[t]
    \centering
    \includegraphics[width=\linewidth]{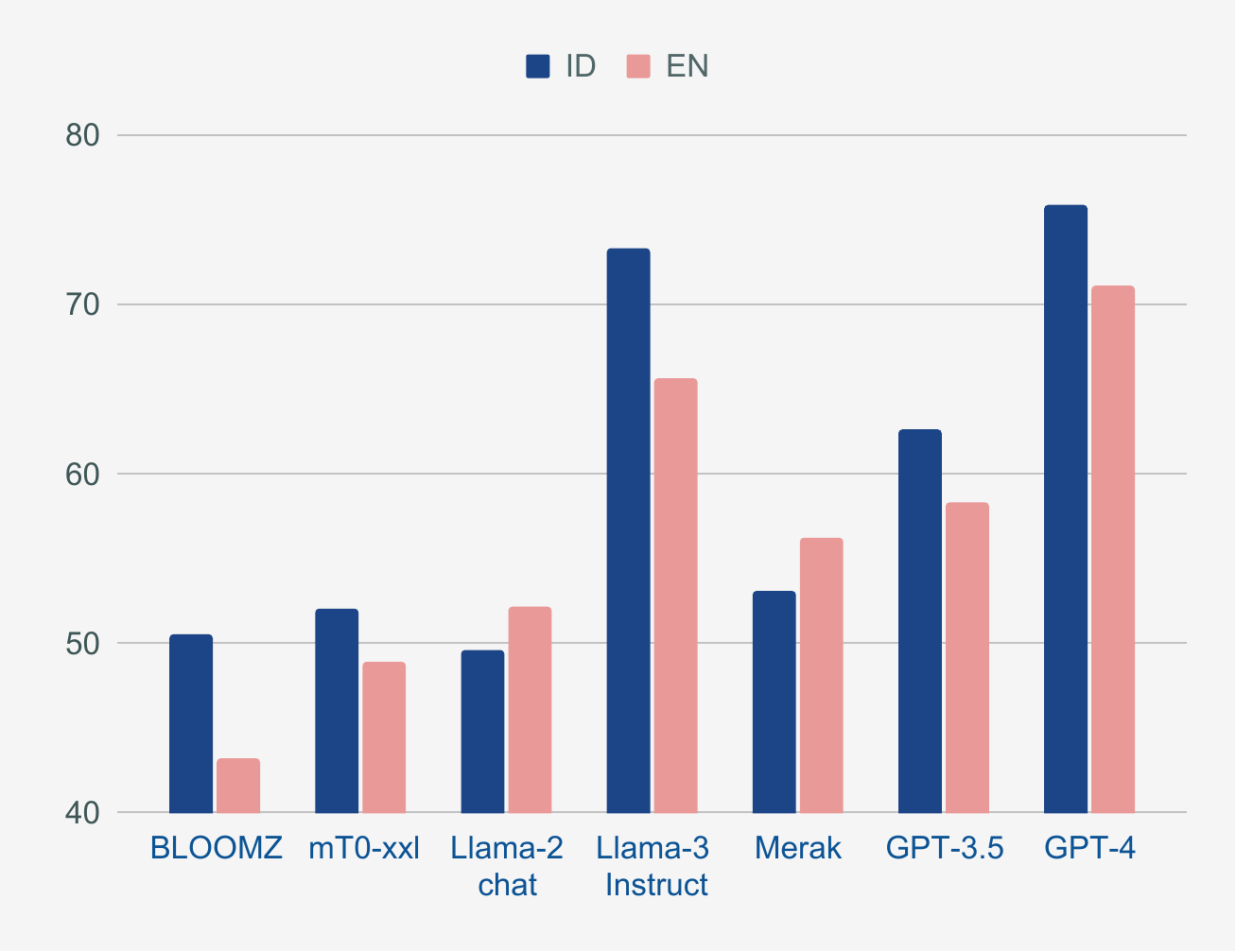} 
    \caption{The accuracy of Indonesian and English translations across BLOOMZ (7B), mT0$_\text{xxl}$ (13B), Llama--2 chat (13B), Llama--3 Instruct (70B), Merak (7B), GPT--3.5, and GPT--4.}
    \label{fig:id_en}
\end{figure}

\section{Discussion}

A recent study \cite{wang-etal-2024-answer-c} has demonstrated that evaluating language models using token probabilities in multiple-choice question types does not align well with the generated text. This discrepancy is reported to be more pronounced in models fine-tuned on conversational or safety data. In response to this issue, we conducted a manual evaluation of 200 random samples (in Section~\ref{sec:analysis}) and found that the performance in the generated text, especially for Merak --- the best open-weight Indonesian-centric language model --- deteriorates. However, this issue is less apparent in larger models such as Llama--3 and GPT--4. Conducting manual evaluation on all data and models is expensive, and we plan to address this issue in future works. This work primarily focuses on introducing a novel dataset constructed for evaluating cultural commonsense reasoning within the Indonesian context, including preliminary evaluation results based on standard methods used in previous studies \cite{OpenAI2023GPT4TR,touvron2023llama,koto-etal-2023-indommlu,li-etal-2024-cmmlu,koto-etal-2024-arabicmmlu}.

\section{Conclusion}

\datasetname{} is a cultural commonsense reasoning dataset encompassing the diversity of Indonesian cultures, spanning from Aceh province in the west to Papua province in the east. Through collaboration with local individuals across eleven provinces and rigorous quality control measures, we introduce \datasetname{} for the purpose of evaluating language models. Our findings reveal that large language models, whether Indonesian-centric or multilingual, demonstrate a limited understanding of Indonesian cultures. Notably, incorporating location as additional context significantly enhances model performance, particularly for GPT--3.5 and GPT--4.

\section*{Limitations}

\datasetname{} is specifically designed to explore the influence of geographical location on cultural commonsense reasoning, with a focus on the present time. It does not consider temporal aspects. Our dataset was created in the year 2023, and we recognize that cultures may evolve over time, as discussed by \citet{mesoudi2016cultural}.

Furthermore, as demonstrated in Section~\ref{sec:analysis}, a significant portion of \datasetname{} comprises cultural elements such as artifacts, norms, and rituals. Symbols, values and beliefs, and language represent smaller proportions, ranging from 4\% to 20\%. We encourage future research to further explore these cultural elements and to expand the geographical coverage beyond the eleven provinces studied in this paper. 

We also acknowledge that our dataset size is relatively small, to cover all the provinces.  However, compared to existing Indonesian datasets that have been manually curated by natives (see COPAL--ID in Table~\ref{tab:dataset_comparison}), ours is significantly larger in terms of both size and regional coverage. Future work may extend the size and the coverage of \datasetname{} to get a more holistic picture of Indonesian cultures. 

\section*{Ethical Considerations}

\datasetname{} is licensed under a Creative Commons Attribution-NonCommercial-ShareAlike 4.0 International License.\footnote{\url{https://creativecommons.org/licenses/by-nc-sa/4.0/}} Our data is intended for academic research and non-commercial purposes. Workers were compensated above the minimum monthly salary in Indonesia and are fully aware that the data will be released to the public. It is important to note that no private or sensitive information of the workers is included in \datasetname{}.

\section*{Acknowledgements}
We acknowledge the thorough feedback and impactful suggestions from the reviewers and the action editor. We also extend our gratitude to the 22 annotators from eleven provinces in Indonesia for their valuable contributions in constructing \datasetname{}. This project was supported by MBZUAI and UI through PUTI-Q2 grant (NKB-1192/UN2.RST/HKP.05.00/2022).

\bibliography{custom,anthology}

\begin{thebibliography}{55}
\expandafter\ifx\csname natexlab\endcsname\relax\def\natexlab#1{#1}\fi

\bibitem[{Aji et~al.(2022)Aji, Winata, Koto, Cahyawijaya, Romadhony, Mahendra, Kurniawan, Moeljadi, Prasojo, Baldwin, Lau, and Ruder}]{aji-etal-2022-one}
Alham~Fikri Aji, Genta~Indra Winata, Fajri Koto, Samuel Cahyawijaya, Ade Romadhony, Rahmad Mahendra, Kemal Kurniawan, David Moeljadi, Radityo~Eko Prasojo, Timothy Baldwin, Jey~Han Lau, and Sebastian Ruder. 2022.
\newblock \href {https://doi.org/10.18653/v1/2022.acl-long.500} {One country, 700+ languages: {NLP} challenges for underrepresented languages and dialects in {I}ndonesia}.
\newblock In \emph{Proceedings of the 60th Annual Meeting of the Association for Computational Linguistics (Volume 1: Long Papers)}, pages 7226--7249, Dublin, Ireland. Association for Computational Linguistics.

\bibitem[{Axtell and Fornwald(1998)}]{axtell1998gestures}
Roger~E Axtell and Mike Fornwald. 1998.
\newblock Gestures: The do's and taboos of body language around the world.
\newblock \emph{Wiley}.

\bibitem[{Bisk et~al.(2020)Bisk, Zellers, Bras, Gao, and Choi}]{DBLP:conf/aaai/BiskZLGC20}
Yonatan Bisk, Rowan Zellers, Ronan~Le Bras, Jianfeng Gao, and Yejin Choi. 2020.
\newblock \href {https://doi.org/10.1609/AAAI.V34I05.6239} {{PIQA:} reasoning about physical commonsense in natural language}.
\newblock In \emph{The Thirty-Fourth {AAAI} Conference on Artificial Intelligence, {AAAI} 2020, The Thirty-Second Innovative Applications of Artificial Intelligence Conference, {IAAI} 2020, The Tenth {AAAI} Symposium on Educational Advances in Artificial Intelligence, {EAAI} 2020}, pages 7432--7439. {AAAI} Press.

\bibitem[{Cahyawijaya et~al.(2021)Cahyawijaya, Winata, Wilie, Vincentio, Li, Kuncoro, Ruder, Lim, Bahar, Khodra, Purwarianti, and Fung}]{cahyawijaya2021indonlg}
Samuel Cahyawijaya, Genta~Indra Winata, Bryan Wilie, Karissa Vincentio, Xiaohong Li, Adhiguna Kuncoro, Sebastian Ruder, Zhi~Yuan Lim, Syafri Bahar, Masayu Khodra, Ayu Purwarianti, and Pascale Fung. 2021.
\newblock {IndoNLG}: Benchmark and resources for evaluating {Indonesian} natural language generation.
\newblock In \emph{Proceedings of the 2021 Conference on Empirical Methods in Natural Language Processing}, pages 8875--8898.

\bibitem[{Davison et~al.(2019)Davison, Feldman, and Rush}]{davison-etal-2019-commonsense}
Joe Davison, Joshua Feldman, and Alexander Rush. 2019.
\newblock \href {https://doi.org/10.18653/v1/D19-1109} {Commonsense knowledge mining from pretrained models}.
\newblock In \emph{Proceedings of the 2019 Conference on Empirical Methods in Natural Language Processing and the 9th International Joint Conference on Natural Language Processing (EMNLP-IJCNLP)}, pages 1173--1178, Hong Kong, China. Association for Computational Linguistics.

\bibitem[{Dubey et~al.(2024)Dubey, Jauhri, Pandey, Kadian, Al-Dahle, Letman, Mathur, Schelten, Yang, Fan et~al.}]{dubey2024llama}
Abhimanyu Dubey, Abhinav Jauhri, Abhinav Pandey, Abhishek Kadian, Ahmad Al-Dahle, Aiesha Letman, Akhil Mathur, Alan Schelten, Amy Yang, Angela Fan, et~al. 2024.
\newblock The {Llama} 3 herd of models.
\newblock \emph{arXiv preprint arXiv:2407.21783}.

\bibitem[{Fung et~al.(2023)Fung, Chakrabarty, Guo, Rambow, Muresan, and Ji}]{fung-etal-2023-normsage}
Yi~Fung, Tuhin Chakrabarty, Hao Guo, Owen Rambow, Smaranda Muresan, and Heng Ji. 2023.
\newblock \href {https://doi.org/10.18653/v1/2023.emnlp-main.941} {{NORMSAGE}: Multi-lingual multi-cultural norm discovery from conversations on-the-fly}.
\newblock In \emph{Proceedings of the 2023 Conference on Empirical Methods in Natural Language Processing}, pages 15217--15230, Singapore. Association for Computational Linguistics.

\bibitem[{Fung et~al.(2024)Fung, Zhao, Doo, Sun, and Ji}]{fung2024massively}
Yi~Fung, Ruining Zhao, Jae Doo, Chenkai Sun, and Heng Ji. 2024.
\newblock Massively multi-cultural knowledge acquisition \& lm benchmarking.
\newblock \emph{arXiv preprint arXiv:2402.09369}.

\bibitem[{Giddens and Sutton(2021)}]{giddens2021essential}
Anthony Giddens and Philip~W Sutton. 2021.
\newblock \emph{Essential Concepts in Sociology}.
\newblock John Wiley \& Sons.

\bibitem[{Hershcovich et~al.(2022)Hershcovich, Frank, Lent, de~Lhoneux, Abdou, Brandl, Bugliarello, Cabello~Piqueras, Chalkidis, Cui, Fierro, Margatina, Rust, and S{\o}gaard}]{hershcovich-etal-2022-challenges}
Daniel Hershcovich, Stella Frank, Heather Lent, Miryam de~Lhoneux, Mostafa Abdou, Stephanie Brandl, Emanuele Bugliarello, Laura Cabello~Piqueras, Ilias Chalkidis, Ruixiang Cui, Constanza Fierro, Katerina Margatina, Phillip Rust, and Anders S{\o}gaard. 2022.
\newblock \href {https://doi.org/10.18653/v1/2022.acl-long.482} {Challenges and strategies in cross-cultural {NLP}}.
\newblock In \emph{Proceedings of the 60th Annual Meeting of the Association for Computational Linguistics (Volume 1: Long Papers)}, pages 6997--7013, Dublin, Ireland. Association for Computational Linguistics.

\bibitem[{Huang and Chang(2023)}]{huang-chang-2023-towards}
Jie Huang and Kevin Chen-Chuan Chang. 2023.
\newblock \href {https://doi.org/10.18653/v1/2023.findings-acl.67} {Towards reasoning in large language models: A survey}.
\newblock In \emph{Findings of the Association for Computational Linguistics: ACL 2023}, pages 1049--1065, Toronto, Canada. Association for Computational Linguistics.

\bibitem[{Ichsan(2023)}]{Merak}
Muhammad Ichsan. 2023.
\newblock Merak-7b: The {LLM} for {Bahasa Indonesia}.
\newblock \emph{Hugging Face Repository}.

\bibitem[{Koto et~al.(2023)Koto, Aisyah, Li, and Baldwin}]{koto-etal-2023-indommlu}
Fajri Koto, Nurul Aisyah, Haonan Li, and Timothy Baldwin. 2023.
\newblock Large language models only pass primary school exams in {I}ndonesia: A comprehensive test on {I}ndo{MMLU}.
\newblock In \emph{Proceedings of the 2023 Conference on Empirical Methods in Natural Language Processing (EMNLP)}, Singapore. Association for Computational Linguistics.

\bibitem[{Koto et~al.(2022)Koto, Baldwin, and Lau}]{koto-etal-2022-cloze}
Fajri Koto, Timothy Baldwin, and Jey~Han Lau. 2022.
\newblock \href {https://doi.org/10.18653/v1/2022.csrr-1.2} {Cloze evaluation for deeper understanding of commonsense stories in {I}ndonesian}.
\newblock In \emph{Proceedings of the First Workshop on Commonsense Representation and Reasoning (CSRR 2022)}, pages 8--16, Dublin, Ireland. Association for Computational Linguistics.

\bibitem[{Koto et~al.(2024)Koto, Li, Shatnawi, Doughman, Sadallah, Alraeesi, Almubarak, Alyafeai, Sengupta, Shehata, Habash, Nakov, and Baldwin}]{koto-etal-2024-arabicmmlu}
Fajri Koto, Haonan Li, Sara Shatnawi, Jad Doughman, Abdelrahman Sadallah, Aisha Alraeesi, Khalid Almubarak, Zaid Alyafeai, Neha Sengupta, Shady Shehata, Nizar Habash, Preslav Nakov, and Timothy Baldwin. 2024.
\newblock \href {https://aclanthology.org/2024.findings-acl.334} {{A}rabic{MMLU}: Assessing massive multitask language understanding in {A}rabic}.
\newblock In \emph{Findings of the Association for Computational Linguistics ACL 2024}, pages 5622--5640, Bangkok, Thailand and virtual meeting. Association for Computational Linguistics.

\bibitem[{Levesque et~al.(2012)Levesque, Davis, and Morgenstern}]{levesque2012winograd}
Hector Levesque, Ernest Davis, and Leora Morgenstern. 2012.
\newblock The winograd schema challenge.
\newblock In \emph{Thirteenth international conference on the principles of knowledge representation and reasoning}.

\bibitem[{Li et~al.(2023)Li, Koto, Wu, Aji, and Baldwin}]{li2023bactrian}
Haonan Li, Fajri Koto, Minghao Wu, Alham~Fikri Aji, and Timothy Baldwin. 2023.
\newblock Bactrian-{X}: A multilingual replicable instruction-following model with low-rank adaptation.
\newblock \emph{arXiv preprint arXiv:2305.15011}.

\bibitem[{Li et~al.(2024)Li, Zhang, Koto, Yang, Zhao, Gong, Duan, and Baldwin}]{li-etal-2024-cmmlu}
Haonan Li, Yixuan Zhang, Fajri Koto, Yifei Yang, Hai Zhao, Yeyun Gong, Nan Duan, and Timothy Baldwin. 2024.
\newblock \href {https://aclanthology.org/2024.findings-acl.671} {{CMMLU}: Measuring massive multitask language understanding in {C}hinese}.
\newblock In \emph{Findings of the Association for Computational Linguistics ACL 2024}, pages 11260--11285, Bangkok, Thailand and virtual meeting. Association for Computational Linguistics.

\bibitem[{Lin et~al.(2020)Lin, Lee, Khanna, and Ren}]{lin-etal-2020-birds}
Bill~Yuchen Lin, Seyeon Lee, Rahul Khanna, and Xiang Ren. 2020.
\newblock \href {https://doi.org/10.18653/v1/2020.emnlp-main.557} {{B}irds have four legs?! {N}umer{S}ense: {P}robing {N}umerical {C}ommonsense {K}nowledge of {P}re-{T}rained {L}anguage {M}odels}.
\newblock In \emph{Proceedings of the 2020 Conference on Empirical Methods in Natural Language Processing (EMNLP)}, pages 6862--6868, Online. Association for Computational Linguistics.

\bibitem[{Lin et~al.(2022)Lin, Mihaylov, Artetxe, Wang, Chen, Simig, Ott, Goyal, Bhosale, Du, Pasunuru, Shleifer, Koura, Chaudhary, O{'}Horo, Wang, Zettlemoyer, Kozareva, Diab, Stoyanov, and Li}]{lin-etal-2022-shot}
Xi~Victoria Lin, Todor Mihaylov, Mikel Artetxe, Tianlu Wang, Shuohui Chen, Daniel Simig, Myle Ott, Naman Goyal, Shruti Bhosale, Jingfei Du, Ramakanth Pasunuru, Sam Shleifer, Punit~Singh Koura, Vishrav Chaudhary, Brian O{'}Horo, Jeff Wang, Luke Zettlemoyer, Zornitsa Kozareva, Mona Diab, Veselin Stoyanov, and Xian Li. 2022.
\newblock \href {https://aclanthology.org/2022.emnlp-main.616} {Few-shot learning with multilingual generative language models}.
\newblock In \emph{Proceedings of the 2022 Conference on Empirical Methods in Natural Language Processing}, pages 9019--9052, Abu Dhabi, United Arab Emirates. Association for Computational Linguistics.

\bibitem[{Liu et~al.(2024)Liu, Koto, Baldwin, and Gurevych}]{liu-etal-2024-multilingual}
Chen Liu, Fajri Koto, Timothy Baldwin, and Iryna Gurevych. 2024.
\newblock \href {https://aclanthology.org/2024.naacl-long.112} {Are multilingual {LLM}s culturally-diverse reasoners? an investigation into multicultural proverbs and sayings}.
\newblock In \emph{Proceedings of the 2024 Conference of the North American Chapter of the Association for Computational Linguistics: Human Language Technologies (Volume 1: Long Papers)}, pages 2016--2039, Mexico City, Mexico. Association for Computational Linguistics.

\bibitem[{Liu et~al.(2023)Liu, Qiao, Neiswanger, Wang, Tan, Tao, Li, Wang, Sun, Pangarkar, Fan, Gu, Miller, Zhuang, He, Li, Koto, Tang, Ranjan, Shen, Ren, Iriondo, Mu, Hu, Schulze, Nakov, Baldwin, and Xing}]{liu2023llm360}
Zhengzhong Liu, Aurick Qiao, Willie Neiswanger, Hongyi Wang, Bowen Tan, Tianhua Tao, Junbo Li, Yuqi Wang, Suqi Sun, Omkar Pangarkar, Richard Fan, Yi~Gu, Victor Miller, Yonghao Zhuang, Guowei He, Haonan Li, Fajri Koto, Liping Tang, Nikhil Ranjan, Zhiqiang Shen, Xuguang Ren, Roberto Iriondo, Cun Mu, Zhiting Hu, Mark Schulze, Preslav Nakov, Timothy Baldwin, and Eric~P. Xing. 2023.
\newblock {LLM360}: Towards fully transparent open-source {LLMs}.
\newblock \emph{arXiv preprint arXiv:2312.06550}.

\bibitem[{Macionis(2012)}]{macionis2012sociology}
John~J. Macionis. 2012.
\newblock \emph{Sociology: Fourteenth Edition}.
\newblock Pearson.

\bibitem[{Madaan et~al.(2022)Madaan, Zhou, Alon, Yang, and Neubig}]{madaan-etal-2022-language}
Aman Madaan, Shuyan Zhou, Uri Alon, Yiming Yang, and Graham Neubig. 2022.
\newblock \href {https://aclanthology.org/2022.emnlp-main.90} {Language models of code are few-shot commonsense learners}.
\newblock In \emph{Proceedings of the 2022 Conference on Empirical Methods in Natural Language Processing}, pages 1384--1403, Abu Dhabi, United Arab Emirates. Association for Computational Linguistics.

\bibitem[{Mahendra et~al.(2021)Mahendra, Aji, Louvan, Rahman, and Vania}]{mahendra-etal-2021-indonli}
Rahmad Mahendra, Alham~Fikri Aji, Samuel Louvan, Fahrurrozi Rahman, and Clara Vania. 2021.
\newblock \href {https://doi.org/10.18653/v1/2021.emnlp-main.821} {{I}ndo{NLI}: A natural language inference dataset for {I}ndonesian}.
\newblock In \emph{Proceedings of the 2021 Conference on Empirical Methods in Natural Language Processing}, pages 10511--10527, Online and Punta Cana, Dominican Republic. Association for Computational Linguistics.

\bibitem[{Mesoudi(2016)}]{mesoudi2016cultural}
Alex Mesoudi. 2016.
\newblock Cultural evolution: Integrating psychology, evolution and culture.
\newblock \emph{Current Opinion in Psychology}, 7:17--22.

\bibitem[{Mostafazadeh et~al.(2016)Mostafazadeh, Chambers, He, Parikh, Batra, Vanderwende, Kohli, and Allen}]{mostafazadeh-etal-2016-corpus}
Nasrin Mostafazadeh, Nathanael Chambers, Xiaodong He, Devi Parikh, Dhruv Batra, Lucy Vanderwende, Pushmeet Kohli, and James Allen. 2016.
\newblock \href {https://doi.org/10.18653/v1/N16-1098} {A corpus and cloze evaluation for deeper understanding of commonsense stories}.
\newblock In \emph{Proceedings of the 2016 Conference of the North {A}merican Chapter of the Association for Computational Linguistics: Human Language Technologies}, pages 839--849, San Diego, California. Association for Computational Linguistics.

\bibitem[{Muennighoff et~al.(2023)Muennighoff, Wang, Sutawika, Roberts, Biderman, Le~Scao, Bari, Shen, Yong, Schoelkopf, Tang, Radev, Aji, Almubarak, Albanie, Alyafeai, Webson, Raff, and Raffel}]{muennighoff-etal-2023-crosslingual}
Niklas Muennighoff, Thomas Wang, Lintang Sutawika, Adam Roberts, Stella Biderman, Teven Le~Scao, M~Saiful Bari, Sheng Shen, Zheng~Xin Yong, Hailey Schoelkopf, Xiangru Tang, Dragomir Radev, Alham~Fikri Aji, Khalid Almubarak, Samuel Albanie, Zaid Alyafeai, Albert Webson, Edward Raff, and Colin Raffel. 2023.
\newblock \href {https://doi.org/10.18653/v1/2023.acl-long.891} {Crosslingual generalization through multitask finetuning}.
\newblock In \emph{Proceedings of the 61st Annual Meeting of the Association for Computational Linguistics (Volume 1: Long Papers)}, pages 15991--16111, Toronto, Canada. Association for Computational Linguistics.

\bibitem[{Myung et~al.(2024)Myung, Lee, Zhou, Jin, Putri, Antypas, Borkakoty, Kim, Perez-Almendros, Ayele, Gutiérrez-Basulto, Ibáñez-García, Lee, Muhammad, Park, Rzayev, White, Yimam, Pilehvar, Ousidhoum, Camacho-Collados, and Oh}]{myung2024blend}
Junho Myung, Nayeon Lee, Yi~Zhou, Jiho Jin, Rifki~Afina Putri, Dimosthenis Antypas, Hsuvas Borkakoty, Eunsu Kim, Carla Perez-Almendros, Abinew~Ali Ayele, Víctor Gutiérrez-Basulto, Yazmín Ibáñez-García, Hwaran Lee, Shamsuddeen~Hassan Muhammad, Kiwoong Park, Anar~Sabuhi Rzayev, Nina White, Seid~Muhie Yimam, Mohammad~Taher Pilehvar, Nedjma Ousidhoum, Jose Camacho-Collados, and Alice Oh. 2024.
\newblock {BLEnD}: A benchmark for {LLMs} on everyday knowledge in diverse cultures and languages.
\newblock \emph{arXiv preprint arXiv:2406.09948}.

\bibitem[{Nguyen et~al.(2023)Nguyen, Razniewski, Varde, and Weikum}]{nguyen2023extracting}
Tuan-Phong Nguyen, Simon Razniewski, Aparna Varde, and Gerhard Weikum. 2023.
\newblock Extracting cultural commonsense knowledge at scale.
\newblock In \emph{Proceedings of the ACM Web Conference 2023}, pages 1907--1917.

\bibitem[{Nguyen et~al.(2024)Nguyen, Zhang, Li, Aljunied, Hu, Shen, Chia, Li, Wang, Tan, Cheng, Chen, Deng, Yang, Liu, Zhang, and Bing}]{nguyen-etal-2024-seallms}
Xuan-Phi Nguyen, Wenxuan Zhang, Xin Li, Mahani Aljunied, Zhiqiang Hu, Chenhui Shen, Yew~Ken Chia, Xingxuan Li, Jianyu Wang, Qingyu Tan, Liying Cheng, Guanzheng Chen, Yue Deng, Sen Yang, Chaoqun Liu, Hang Zhang, and Lidong Bing. 2024.
\newblock \href {https://aclanthology.org/2024.acl-demos.28} {{S}ea{LLM}s - large language models for {S}outheast {A}sia}.
\newblock In \emph{Proceedings of the 62nd Annual Meeting of the Association for Computational Linguistics (Volume 3: System Demonstrations)}, pages 294--304, Bangkok, Thailand. Association for Computational Linguistics.

\bibitem[{OpenAI(2023)}]{OpenAI2023GPT4TR}
OpenAI. 2023.
\newblock {GPT-4} technical report.
\newblock \emph{ArXiv}, abs/2303.08774.

\bibitem[{Ouyang et~al.(2022)Ouyang, Wu, Jiang, Almeida, Wainwright, Mishkin, Zhang, Agarwal, Slama, Ray, Schulman, Hilton, Kelton, Miller, Simens, Askell, Welinder, Christiano, Leike, and Lowe}]{ouyang2022training}
Long Ouyang, Jeffrey Wu, Xu~Jiang, Diogo Almeida, Carroll Wainwright, Pamela Mishkin, Chong Zhang, Sandhini Agarwal, Katarina Slama, Alex Ray, John Schulman, Jacob Hilton, Fraser Kelton, Luke Miller, Maddie Simens, Amanda Askell, Peter Welinder, Paul~F. Christiano, Jan Leike, and Ryan Lowe. 2022.
\newblock Training language models to follow instructions with human feedback.
\newblock \emph{Advances in Neural Information Processing Systems}, 35:27730--27744.

\bibitem[{Owen et~al.(2024)Owen, Tripathi, Kumar, and Ahmed}]{komodo}
Louis Owen, Vishesh Tripathi, Abhay Kumar, and Biddwan Ahmed. 2024.
\newblock Komodo: A linguistic expedition into {Indonesia's} regional languages.
\newblock \emph{arXiv preprint arXiv:2403.09362}.

\bibitem[{Ponti et~al.(2020)Ponti, Glava{\v{s}}, Majewska, Liu, Vuli{\'c}, and Korhonen}]{ponti-etal-2020-xcopa}
Edoardo~Maria Ponti, Goran Glava{\v{s}}, Olga Majewska, Qianchu Liu, Ivan Vuli{\'c}, and Anna Korhonen. 2020.
\newblock \href {https://doi.org/10.18653/v1/2020.emnlp-main.185} {{XCOPA}: A multilingual dataset for causal commonsense reasoning}.
\newblock In \emph{Proceedings of the 2020 Conference on Empirical Methods in Natural Language Processing (EMNLP)}, pages 2362--2376, Online. Association for Computational Linguistics.

\bibitem[{Putra et~al.(2019)Putra, Mahendra, and Darari}]{DBLP:conf/wims/PutraMD19}
Hadi~Syah Putra, Rahmad Mahendra, and Fariz Darari. 2019.
\newblock \href {https://doi.org/10.1145/3326467.3326487} {Budayakb: Extraction of cultural heritage entities from heterogeneous formats}.
\newblock In \emph{Proceedings of the 9th International Conference on Web Intelligence, Mining and Semantics, {WIMS} 2019}, pages 6:1--6:9. {ACM}.

\bibitem[{Putri et~al.(2024)Putri, Ghifari~Haznitrama, Adhista, and Oh}]{afina2024can}
Rifki~Afina Putri, Faiz Ghifari~Haznitrama, Dea Adhista, and Alice Oh. 2024.
\newblock Can {LLM} generate culturally relevant commonsense {QA} data? case study in {Indonesian} and {Sundanese}.
\newblock \emph{arXiv e-prints}, pages arXiv--2402.

\bibitem[{Qin et~al.(2021)Qin, Gupta, Upadhyay, He, Choi, and Faruqui}]{qin-etal-2021-timedial}
Lianhui Qin, Aditya Gupta, Shyam Upadhyay, Luheng He, Yejin Choi, and Manaal Faruqui. 2021.
\newblock \href {https://doi.org/10.18653/v1/2021.acl-long.549} {{TIMEDIAL}: Temporal commonsense reasoning in dialog}.
\newblock In \emph{Proceedings of the 59th Annual Meeting of the Association for Computational Linguistics and the 11th International Joint Conference on Natural Language Processing (Volume 1: Long Papers)}, pages 7066--7076, Online. Association for Computational Linguistics.

\bibitem[{Roemmele et~al.(2011)Roemmele, Bejan, and Gordon}]{roemmele2011choice}
Melissa Roemmele, Cosmin~Adrian Bejan, and Andrew~S Gordon. 2011.
\newblock Choice of plausible alternatives: An evaluation of commonsense causal reasoning.
\newblock In \emph{2011 AAAI Spring Symposium Series}.

\bibitem[{Sakaguchi et~al.(2021)Sakaguchi, Bras, Bhagavatula, and Choi}]{sakaguchi2021winogrande}
Keisuke Sakaguchi, Ronan~Le Bras, Chandra Bhagavatula, and Yejin Choi. 2021.
\newblock Winogrande: An adversarial {Winograd} schema challenge at scale.
\newblock \emph{Communications of the ACM}, 64(9):99--106.

\bibitem[{Sap et~al.(2019)Sap, Rashkin, Chen, Le~Bras, and Choi}]{sap-etal-2019-social}
Maarten Sap, Hannah Rashkin, Derek Chen, Ronan Le~Bras, and Yejin Choi. 2019.
\newblock \href {https://doi.org/10.18653/v1/D19-1454} {Social {IQ}a: Commonsense reasoning about social interactions}.
\newblock In \emph{Proceedings of the 2019 Conference on Empirical Methods in Natural Language Processing and the 9th International Joint Conference on Natural Language Processing (EMNLP-IJCNLP)}, pages 4463--4473, Hong Kong, China. Association for Computational Linguistics.

\bibitem[{Sengupta et~al.(2023)Sengupta, Sahu, Jia, Katipomu, Li, Koto, Marshall, Gosal, Liu, Chen, Afzal, Kamboj, Pandit, Pal, Pradhan, Mujahid, Baali, Han, Bsharat, Aji, Shen, Liu, Vassilieva, Hestness, Hock, Feldman, Lee, Jackson, Ren, Nakov, Baldwin, and Xing}]{sengupta2023jais}
Neha Sengupta, Sunil~Kumar Sahu, Bokang Jia, Satheesh Katipomu, Haonan Li, Fajri Koto, William Marshall, Gurpreet Gosal, Cynthia Liu, Zhiming Chen, Osama~Mohammed Afzal, Samta Kamboj, Onkar Pandit, Rahul Pal, Lalit Pradhan, Zain~Muhammad Mujahid, Massa Baali, Xudong Han, Sondos~Mahmoud Bsharat, Alham~Fikri Aji, Zhiqiang Shen, Zhengzhong Liu, Natalia Vassilieva, Joel Hestness, Andy Hock, Andrew Feldman, Jonathan Lee, Andrew Jackson, Hector~Xuguang Ren, Preslav Nakov, Timothy Baldwin, and Eric Xing. 2023.
\newblock Jais and {Jais-chat}: {Arabic}-centric foundation and instruction-tuned open generative large language models.
\newblock \emph{arXiv preprint arXiv:2308.16149}.

\bibitem[{Shwartz(2022)}]{shwartz-2022-good}
Vered Shwartz. 2022.
\newblock \href {https://doi.org/10.18653/v1/2022.findings-acl.224} {Good night at 4 pm?! time expressions in different cultures}.
\newblock In \emph{Findings of the Association for Computational Linguistics: ACL 2022}, pages 2842--2853, Dublin, Ireland. Association for Computational Linguistics.

\bibitem[{Singapore(2023)}]{sea_lion_2023}
AI~Singapore. 2023.
\newblock {SEA-LION} ({Southeast Asian} languages in one network): A family of large language models for {Southeast Asia}.
\newblock \url{https://github.com/aisingapore/sealion}.

\bibitem[{Thomas(1983)}]{thomas1983cross}
Jenny Thomas. 1983.
\newblock Cross-cultural pragmatic failure.
\newblock \emph{Applied linguistics}, 4(2):91--112.

\bibitem[{Touvron et~al.(2023)Touvron, Martin, Stone, Albert, Almahairi, Babaei, Bashlykov, Batra, Bhargava, Bhosale et~al.}]{touvron2023llama}
Hugo Touvron, Louis Martin, Kevin Stone, Peter Albert, Amjad Almahairi, Yasmine Babaei, Nikolay Bashlykov, Soumya Batra, Prajjwal Bhargava, Shruti Bhosale, et~al. 2023.
\newblock Llama 2: Open foundation and fine-tuned chat models.
\newblock \emph{arXiv preprint arXiv:2307.09288}.

\bibitem[{Wang et~al.(2024)Wang, Ma, Hu, Weber-Genzel, R{\"o}ttger, Kreuter, Hovy, and Plank}]{wang-etal-2024-answer-c}
Xinpeng Wang, Bolei Ma, Chengzhi Hu, Leon Weber-Genzel, Paul R{\"o}ttger, Frauke Kreuter, Dirk Hovy, and Barbara Plank. 2024.
\newblock \href {https://aclanthology.org/2024.findings-acl.441} {{``}{My} answer is {C}{''}: First-token probabilities do not match text answers in instruction-tuned language models}.
\newblock In \emph{Findings of the Association for Computational Linguistics ACL 2024}, pages 7407--7416, Bangkok, Thailand and virtual meeting. Association for Computational Linguistics.

\bibitem[{Wibowo et~al.(2024)Wibowo, Fuadi, Nityasya, Prasojo, and Aji}]{wibowo-etal-2024-copal}
Haryo Wibowo, Erland Fuadi, Made Nityasya, Radityo~Eko Prasojo, and Alham Aji. 2024.
\newblock \href {https://aclanthology.org/2024.naacl-long.77} {{COPAL}-{ID}: {I}ndonesian language reasoning with local culture and nuances}.
\newblock In \emph{Proceedings of the 2024 Conference of the North American Chapter of the Association for Computational Linguistics: Human Language Technologies (Volume 1: Long Papers)}, pages 1404--1422, Mexico City, Mexico. Association for Computational Linguistics.

\bibitem[{Williams(2014)}]{williams2014keywords}
Raymond Williams. 2014.
\newblock \emph{Keywords: A vocabulary of culture and society}.
\newblock Oxford university press.

\bibitem[{Winata et~al.(2023)Winata, Aji, Cahyawijaya, Mahendra, Koto, Romadhony, Kurniawan, Moeljadi, Prasojo, Fung, Baldwin, Lau, Sennrich, and Ruder}]{winata-etal-2023-nusax}
Genta~Indra Winata, Alham~Fikri Aji, Samuel Cahyawijaya, Rahmad Mahendra, Fajri Koto, Ade Romadhony, Kemal Kurniawan, David Moeljadi, Radityo~Eko Prasojo, Pascale Fung, Timothy Baldwin, Jey~Han Lau, Rico Sennrich, and Sebastian Ruder. 2023.
\newblock \href {https://aclanthology.org/2023.eacl-main.57} {{N}usa{X}: Multilingual parallel sentiment dataset for 10 {I}ndonesian local languages}.
\newblock In \emph{Proceedings of the 17th Conference of the European Chapter of the Association for Computational Linguistics}, pages 815--834, Dubrovnik, Croatia. Association for Computational Linguistics.

\bibitem[{Wolf et~al.(2020)Wolf, Debut, Sanh, Chaumond, Delangue, Moi, Cistac, Rault, Louf, Funtowicz, Davison, Shleifer, von Platen, Ma, Jernite, Plu, Xu, Le~Scao, Gugger, Drame, Lhoest, and Rush}]{wolf-etal-2020-transformers}
Thomas Wolf, Lysandre Debut, Victor Sanh, Julien Chaumond, Clement Delangue, Anthony Moi, Pierric Cistac, Tim Rault, Remi Louf, Morgan Funtowicz, Joe Davison, Sam Shleifer, Patrick von Platen, Clara Ma, Yacine Jernite, Julien Plu, Canwen Xu, Teven Le~Scao, Sylvain Gugger, Mariama Drame, Quentin Lhoest, and Alexander Rush. 2020.
\newblock \href {https://doi.org/10.18653/v1/2020.emnlp-demos.6} {Transformers: State-of-the-art natural language processing}.
\newblock In \emph{Proceedings of the 2020 Conference on Empirical Methods in Natural Language Processing: System Demonstrations}, pages 38--45, Online. Association for Computational Linguistics.

\bibitem[{Yin et~al.(2022)Yin, Bansal, Monajatipoor, Li, and Chang}]{yin-etal-2022-geomlama}
Da~Yin, Hritik Bansal, Masoud Monajatipoor, Liunian~Harold Li, and Kai-Wei Chang. 2022.
\newblock \href {https://aclanthology.org/2022.emnlp-main.132} {{G}eo{MLAMA}: Geo-diverse commonsense probing on multilingual pre-trained language models}.
\newblock In \emph{Proceedings of the 2022 Conference on Empirical Methods in Natural Language Processing}, pages 2039--2055, Abu Dhabi, United Arab Emirates. Association for Computational Linguistics.

\bibitem[{Zarbaliyev(2017)}]{zarbaliyev2017multiculturalism}
Habib Zarbaliyev. 2017.
\newblock Multiculturalism in globalization era: History and challenge for {Indonesia}.
\newblock \emph{Journal of Social Studies (JSS)}, 13(1):1--16.

\bibitem[{Zellers et~al.(2019)Zellers, Holtzman, Bisk, Farhadi, and Choi}]{zellers-etal-2019-hellaswag}
Rowan Zellers, Ari Holtzman, Yonatan Bisk, Ali Farhadi, and Yejin Choi. 2019.
\newblock \href {https://doi.org/10.18653/v1/P19-1472} {{H}ella{S}wag: Can a machine really finish your sentence?}
\newblock In \emph{Proceedings of the 57th Annual Meeting of the Association for Computational Linguistics}, pages 4791--4800, Florence, Italy. Association for Computational Linguistics.

\bibitem[{Ziems et~al.(2023)Ziems, Dwivedi-Yu, Wang, Halevy, and Yang}]{ziems-etal-2023-normbank}
Caleb Ziems, Jane Dwivedi-Yu, Yi-Chia Wang, Alon Halevy, and Diyi Yang. 2023.
\newblock \href {https://doi.org/10.18653/v1/2023.acl-long.429} {{N}orm{B}ank: A knowledge bank of situational social norms}.
\newblock In \emph{Proceedings of the 61st Annual Meeting of the Association for Computational Linguistics (Volume 1: Long Papers)}, pages 7756--7776, Toronto, Canada. Association for Computational Linguistics.

\end{thebibliography}
\bibliographystyle{acl_natbib}

\appendix


\section{Model Details}
\label{ap:model}

\begin{table}[ht!]
    \centering
    \resizebox{\linewidth}{!}{
        \begin{tabular}{lr}
        \toprule
        \textbf{Models (\#parameters)} & \textbf{Source} \\
        \midrule
        BLOOMZ (560M)  & \texttt{bigscience/bloomz-560m} \\
        BLOOMZ (1.1B)  & \texttt{bigscience/bloomz-1b1} \\
        BLOOMZ (1.7B)  & \texttt{bigscience/bloomz-1b7} \\
        BLOOMZ (3B)  & \texttt{bigscience/bloomz-3b} \\
        BLOOMZ (7.1B)  & \texttt{bigscience/bloomz-7b1} \\
        \hdashline 
        mT0$_\text{small}$ (300M)   & \texttt{bigscience/mt0-small} \\
        mT0$_\text{base}$ (580M)  & \texttt{bigscience/mt0-base} \\
        mT0$_\text{large}$ (1.2B)   & \texttt{bigscience/mt0-large} \\
        mT0$_\text{xl}$ (3.7B)   & \texttt{bigscience/mt0-xl} \\
        mT0$_\text{xxl}$ (13B)  & \texttt{bigscience/mt0-xxl} \\
        \hdashline 
        Llama--2 (7B)  & \texttt{meta-llama/Llama--2-7b} \\
        Llama--2 chat (7B)  & \texttt{meta-llama/Llama--2-7b-chat} \\
        Llama--2 (13B)  & \texttt{meta-llama/Llama--2-13b} \\
        Llama--2 chat (13B)  & \texttt{meta-llama/Llama--2-13b-chat} \\
        \hdashline 
        Llama--3 (8B)  & \texttt{meta-llama/Meta-Llama--3-8B} \\
        Llama--3 Instruct (8B)  & \texttt{meta-llama/Meta-Llama--3-8B-Instruct} \\
        Llama--3 (70B)  & \texttt{meta-llama/Meta-Llama--3-70B} \\
        Llama--3-chat (70B)  & \texttt{meta-llama/Meta-Llama--3-70B-Instruct} \\
        \hdashline 
        Bactrian-X$_{\texttt{LLaMa}}$ (7B) & \texttt{MBZUAI/bactrian-x-llama-7b-merged} \\
        Bactrian-X$_{\texttt{LLaMa}}$ (13B) & \texttt{MBZUAI/bactrian-x-llama-13b-merged} \\
        \hdashline 
        IndoBART (132M)  & \texttt{indobenchmark/indobart-v2} \\
        IndoGPT (117M)  & \texttt{indobenchmark/indogpt} \\
        Merak (7B)  & \texttt{Ichsan2895/Merak-7B-v5-PROTOTYPE1} \\
        SeaLLM (7B)  & \texttt{SeaLLMs/SeaLLM-7B-v2} \\
        SEA-LION (7B)  & \texttt{aisingapore/sea-lion-7b} \\
        Komodo (7B)  & \texttt{Yellow-AI-NLP/komodo-7b-base} \\
        \bottomrule
        \end{tabular}
    }
    \caption{With the exception of GPT--3.5 and GPT--4, all the models used in this study were sourced from Huggingface \cite{wolf-etal-2020-transformers}.}
    \label{tab:models}
\end{table}

\end{document}